\documentclass[runningheads]{llncs}

\usepackage{graphicx}
\usepackage{comment}
\usepackage{amsmath,amssymb}
\usepackage{color}
\usepackage{url}
\usepackage{subcaption}
\usepackage[inline]{enumitem}
\usepackage{pifont}
\usepackage{todonotes}
\usepackage[pagebackref=true,breaklinks=true,colorlinks,bookmarks=false]{hyperref}

\newif\ifreview
\reviewfalse

\ifreview
	\usepackage{lineno}

	\linenumbers
\fi

\newcommand{\etal}{\textit{et al.}}

\newcommand{\cmark}{\ding{51}}%
\newcommand{\xmark}{\ding{55}}%

\newcommand{\kppred}{\hat{\mathcal{K}}_{2D,c}}

\newcommand{\kpdpred}{\hat{\mathcal{K}}_{3D}}

\newcommand{\pose}{{\mathcal{P}}_{7D}}
\newcommand{\posepred}{\hat{\mathcal{P}}_{7D}}

\newcommand{\transformk}[1]{\mathcal{T}_{wld} (\mathcal{K}_{o},#1)}
\newcommand{\transformv}[1]{\mathcal{T}_{wld} (\mathcal{V}_{o},#1)}

\begin{document}


\def\SubNumber{74}

\def\GCPRTrack{Fast Review Track}

\title{SurgeoNet: Realtime 3D Pose Estimation of Articulated Surgical Instruments from Stereo Images using a Synthetically-trained Network}

\ifreview
\else
	\titlerunning{SurgeoNet}

	\author{Ahmed Tawfik Aboukhadra\inst{1, 2} \and
        Nadia Robertini\inst{1} \and
        Jameel Malik\inst{3}\and
        Ahmed Elhayek\inst{4}\and
        Gerd Reis\inst{1}\and
        Didier Stricker\inst{1,2}
        }
	
	\authorrunning{AT. Aboukhadra et al.}
	
	\institute{German Research Center for Artificial Intelligence (DFKI), Trippstadter Straße 122, 67663 Kaiserslautern, Germany \\
        \email{\{firstname[\_secondname].lastname\}@dfki.de} \and
        University of Kaiserslautern-Landau (RPTU), Erwin-Schrödinger-Straße 52, 67663 Kaiserslautern, Germany \and
        NUST-SEECS, Islamabad, Pakistan \and
        University of Prince Mugrin (UPM), 42241 Madinah, Saudi Arabia}
\fi

\maketitle              

\begin{abstract}
Surgery monitoring in Mixed Reality (MR) environments has recently received substantial focus due to its importance in image-based decisions, skill assessment, and robot-assisted surgery.
Tracking hands and articulated
surgical instruments is crucial for the success of these applications.
Due to the lack of annotated datasets and the complexity of the task, only a few works have addressed this problem.
In this work, we present SurgeoNet, a real-time neural network pipeline to accurately detect and track surgical instruments from a stereo VR view. Our multi-stage approach is inspired by state-of-the-art neural-network architectural design, like YOLO and Transformers. We demonstrate the generalization capabilities of SurgeoNet in challenging real-world scenarios, achieved solely through training on synthetic data.
The approach can be easily extended to any new set of articulated surgical instruments. 
SurgeoNet's code and data are publicly available\footnote{ 
\url{https://github.com/ATAboukhadra/SurgeoNet}}.




\end{abstract}

\section{Introduction}

Recent advancements in virtual and augmented reality technology have enabled highly immersive gaming, interactive simulation, and virtual experiences, among others.
The use of mixed reality in the medical field is gaining attention, especially in the simulation of medical scenarios for the training of medical personnel outside the laboratory.
To ensure a highly immersive experience, attention goes into realistic user interactions with virtual objects. 
This includes realistic object modeling and pose tracking at low latency. 
The vast majority of the existing approaches in the field of object-tracking, mainly focus on large rigid objects of everyday use, such as books, cans, and cups \cite{hampali2020honnotate,hampali2022keypoint,aboukhadra2023thor}. 
The problem of tracking semi-rigid objects is understudied and has only recently gained attention \cite{fan2023arctic,zhu2023contactart,fan2023arctic,li2020category}. 
Especially in the case of tracking surgical instruments, to the best of our knowledge, there exists little to no advance, due to its intrinsic difficulties \cite{wang2023pov,hein2021towards}. 
Surgical semi-rigid instruments, including thin scissors of various types and forceps or clamps, have strong similarities in shape and appearance. 
On top of it, hand interaction
reduces their visibility, causing typical Computer Vision solutions to struggle to identify or classify them. 
General neural-network-based solutions require tons of finely labeled data for training to succeed at the task. 
However, such data is hard to obtain automatically.

In this work, we introduce SurgeoNet, a real-time solution to the problem of surgical instrument tracking from a stereo (VR) view, that does not rely on a realistic dataset. 
Instead, our approach builds on top of synthetically generated samples of surgical instruments. 

We design our pipeline to infer real-time 7D surgical instruments' pose (3D translation, + 3D rotation + 1D articulation angle) from stereo view and demonstrate its capabilities in detecting, classifying, and tracking instruments in real settings as seen from VR glasses. 
Our method consists of two main components: the first is designed for object detection, classification, and 2D keypoint estimation. 
The second and final part of the pipeline combines the keypoints obtained from left and right stereo-view to infer the corresponding pose. 
Our method is robust to occlusions due to hand interactions and accurately classifies surgical instruments, despite their strong similarities. 
Thanks to its real-time computational performances, our method is suitable for virtual and augmented reality applications, enabling realistic and highly immersive interactions with realistic virtual medical tools. 
The method is easily extendable to a different subset of rigid or semi-rigid medical instruments, currently with at most 1 degree of freedom, and the introduction of new objects in the set is straightforward.

In summary, this work presents SurgeoNet, a new method to accurately reconstruct 7D poses of articulated surgical instruments from stereo view,  with the following key features: 
\begin{enumerate}
    \item Real-time performance, thus suitable for mixed-reality applications;
    \item Reliable classification of surgical instruments of similar shape and appearance under occlusions;
    \item Temporally consistent, jitter-free, tracking.
    \item High generalization capabilities to unseen (real) sequences, despite relying solely on a synthetic dataset.
\end{enumerate}


\section{Related Work}

In our review of related works, we subdivide our problem into three fields: surgical instrument pose estimation, stereo-based object pose estimation, and articulated object pose estimation.

Multiple works studied the surgical instruments' pose estimation problem~\cite{wang2023pov,hein2021towards}. 
Rodrigues~\etal~\cite{rodrigues2022surgical} published a survey of all datasets of surgical instruments. 
Most of those datasets, however, only contain 2D annotations i.e. instrument labels, bounding boxes, 2D keypoints, or at best, segmentation masks. 
An example of recent surgical instruments datasets that contain 2D labels and keypoints is PWISeg~\cite{sun2023pwiseg}.
Hein~\etal~\cite{hein2021towards} proposed a clinical dataset that includes synthetic and real monocular RGB images for hands interacting with a surgical drill along with the 6D annotations of the instrument. 
They also propose a pipeline for hand-object pose estimation, however, it's only focused on a single fully-rigid object i.e. drill. 
In their experiments, they compare the performance of PVNet~\cite{peng2019pvnet} and HandObjectNet~\cite{hasson2020leveraging}.
HMD-EgoPose~\cite{doughty2022hmd} uses an EfficientNet~\cite{tan2020efficientdet} as a backbone to predict the drill pose in the Hein~\etal~dataset. 
In addition, the authors deploy their method on a Microsoft HoloLens $2$ AR headset.
In our experiments section, we finetune our network on the Hein~\etal~real dataset and report pose estimation errors.

POV-Surgery~\cite{wang2023pov} is another work that provides a synthetic dataset that considers temporal dependencies. 
It includes hands wearing stained surgical gloves and interacting with surgical instruments. 
The authors provide a finetuned hand pose estimation model to handle those special-looking hands. 
POV-Surgery focuses only on a small set of completely rigid surgical instruments.

Given our interest in stereo-based vision, we also study the stereo-RGB methods for object pose estimation. 
StereOBJ-1M~\cite{liu2021stereobj} is a large-scale dataset that contains stereo RGB frames of $18$ objects and their 3D pose annotations.
KeyPose~\cite{liu2020keypose} is one of the famous methods meant for stereo-based pose estimation and was evaluated on the StereOBJ-1M dataset.
KeyPose is a neural network that predicts 3D keypoints of rigid transparent objects from stereo input.
One of the key ideas in KeyPose is that they use early fusion in their CNN-based network. 
This means that features from left and right views are merged earlier in the pipeline which improves performance. 

Recently, more attention has been given to articulated object pose estimation~\cite{zhu2023contactart,fan2023arctic,li2020category}.
The ARCTIC dataset~\cite{fan2023arctic} provides a real RGB dataset with full annotations of hands dexterously manipulating articulated objects. 
However, the set of objects provided in those datasets doesn't include surgical instruments and only focuses on everyday objects with varying textures and shapes. 


\section{Method}
\begin{figure}
    \centering
    \includegraphics[width=0.98\linewidth,trim={0 2cm 0 1.5cm},clip]{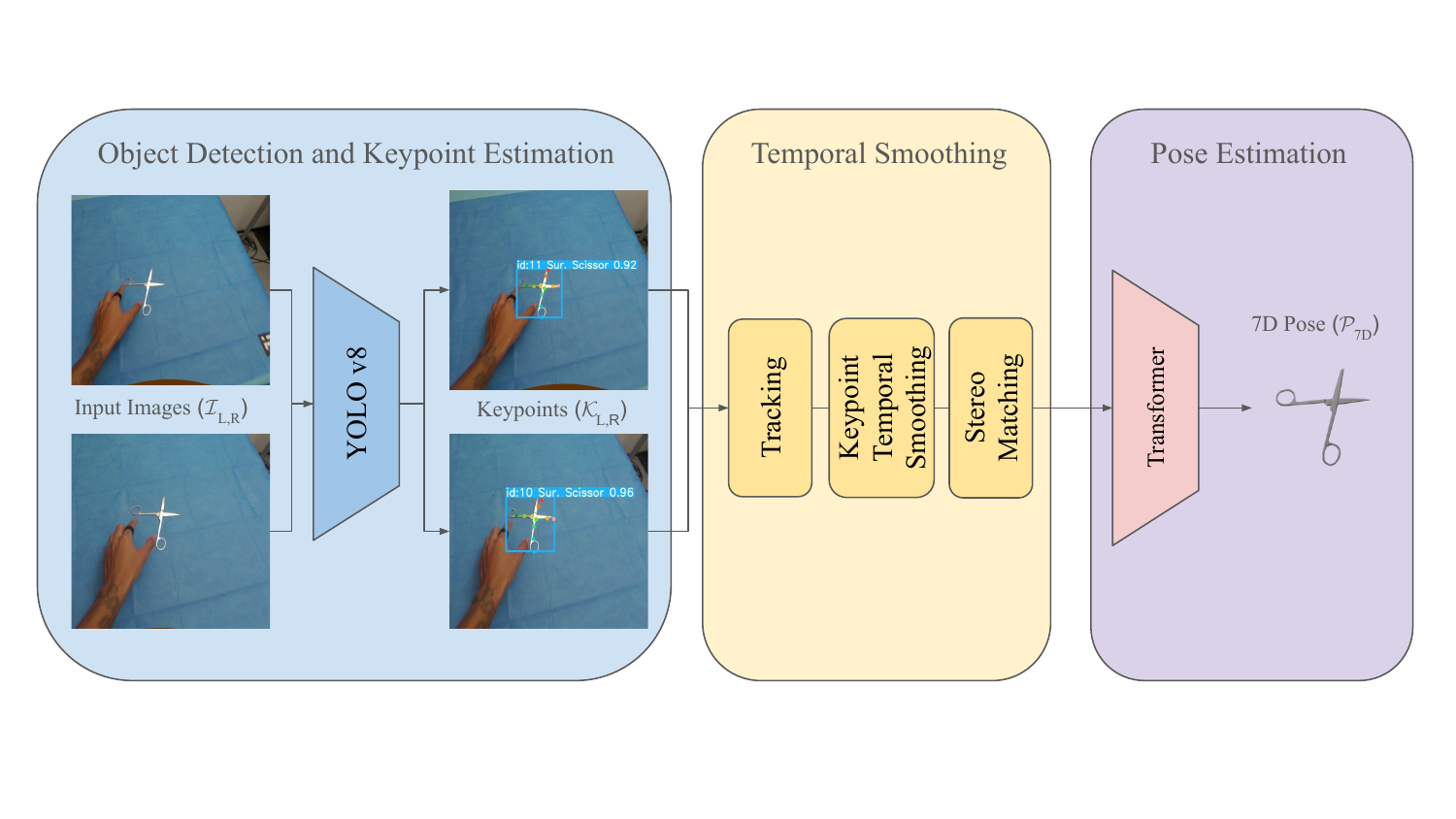}
    \caption{SurgeoNet Architecture.}
    \label{fig:network}
\end{figure}

Given a sequence of calibrated stereo pairs of RGB images, captured from typical VR glasses, mimicking a user's eyes, SurgeoNet estimates the 7D pose of the visible surgical instruments in the camera coordinate system. We propose a neural network pipeline consisting of three stages:
\begin{enumerate*}
    \item Object detection and keypoints estimation.
    \item Tracking and temporal smoothing of keypoints and labels.
    \item 7D Pose Estimation from keypoints and labels.
\end{enumerate*}
Figure \ref{fig:network} shows our selected architecture.
In this section, we describe each network component as well as the synthetic dataset generation process in detail.

\subsubsection{Object Detection and Keypoints Estimation}
The first component of the pipeline detects the surgical instruments present in the views and estimates the corresponding keypoints $\kppred$. 
This stage is implemented using an enhanced version of the YOLO~\cite{redmon2018yolov3} architecture, namely YOLOv8~\cite{yolov8_ultralytics}.
YOLOv8 contains advanced backbone and nick architectures for improved feature extraction.
All YOLO's predictions with a confidence value below $0.7$ are discarded.

\subsubsection{Tracking and temporal smoothing of keypoints and labels}
We use ByteTrack~\cite{zhang2022bytetrack}, a state-of-the-art tracker, to assign a tracking ID to the detected bounding boxes, taking into account their historical positions.
ByteTrack is a MOT algorithm that improves over previous MOT algorithms by associating almost all detections instead of only the high-scoring ones which improves the tracking in case of occlusions. 
ByteTrack predicts the new location of tracks from previous frames using Kalman Fitlers. 
It then uses similarity metrics like IoU and Re-ID to associate new bounding boxes to the tracks. 
This way it can keep unique IDs for the tracks.
Those IDs are used later to apply temporal smoothing to keypoints and reduce jitter. 
For this task we use 1€ Filter~\cite{casiez20121}.
Temporally smoothed keypoints and bounding boxes from the left and right stereo view are finally matched considering the epipolar lines.


\subsubsection{7D Pose Estimation from keypoints and labels}

Taking inspiration from previous work~\cite{zhao2022graformer,zheng20213d}, we designed a Transformer~\cite{vaswani2017attention} network that transforms the stereo 2D keypoints denoted as $\kppred$
of an object to its 7D Pose denoted as $\posepred$ that contains 3D rotation, 3D translation, and 1 articulation value for articulated-like objects as scissors. 
Given that attention layers are permutation invariant i.e. the order of keypoints given to it doesn't change its output.
Therefore, we added a one-hot vector to each keypoint vector to describe its relative position to other keypoints.
Furthermore, to improve the transformer's ability to comprehend which kind of object the keypoints describe, we also append a one-hot encoding that describes the label of the object. 

The network consists of stacked multi-headed attention layers.
the outputs of this network are two-fold:
\begin{enumerate*}
    \item 3D keypoints that correspond to the triangulation of the stereo keypoints $\kpdpred$.
    \item  The 7D object pose $\posepred$ that contains $3$ translation values, $6$ rotation values as described in \cite{zhou2019continuity} that solves the problem of rotation continuity, and $1$ value for articulation angle.
\end{enumerate*}
To train the network, we use the synthetic dataset described in Section \ref{sec:poseDataset}. 
After getting $\posepred$ from the Transformer, it is used in $\mathcal{T}_{wld}$ to transform mesh vertices $\mathcal{V}_{o}$. The loss function used to train this network is:
\begin{equation}
    \mathcal{L}_{p} = \Vert\posepred - \pose\Vert_2 + \Vert\transformv{\posepred} - \transformv{\pose}\Vert_2 +  \Vert\kpdpred - \transformk{\pose}\Vert_2
\end{equation}



\subsection{Synthetic Data}

\subsubsection{Mesh Modeling and Keypoint Selection}
\label{sec:meshes}


To generate synthetic samples of a surgical instrument, we require a 3D model of that object.
This can be acquired from open-source collections or semi-automatic 3D scanning.
To enable modeling of the articulation angle 
for some semi-rigid instruments, e.g. scissors, we manually separate the meshes into their rigid geometric components, e.g. the left and right blades. 
Figure \ref{fig:tools} shows the selected set of surgical instruments.
Out of those 3D models, we choose a set of keypoints that will be used later for building and training our deep neural network approach.
\subsubsection{Synthetic RGB Dataset}
\begin{figure}
    \centering
    \begin{subfigure}{0.41\textwidth}
        \includegraphics[width=0.99\linewidth]{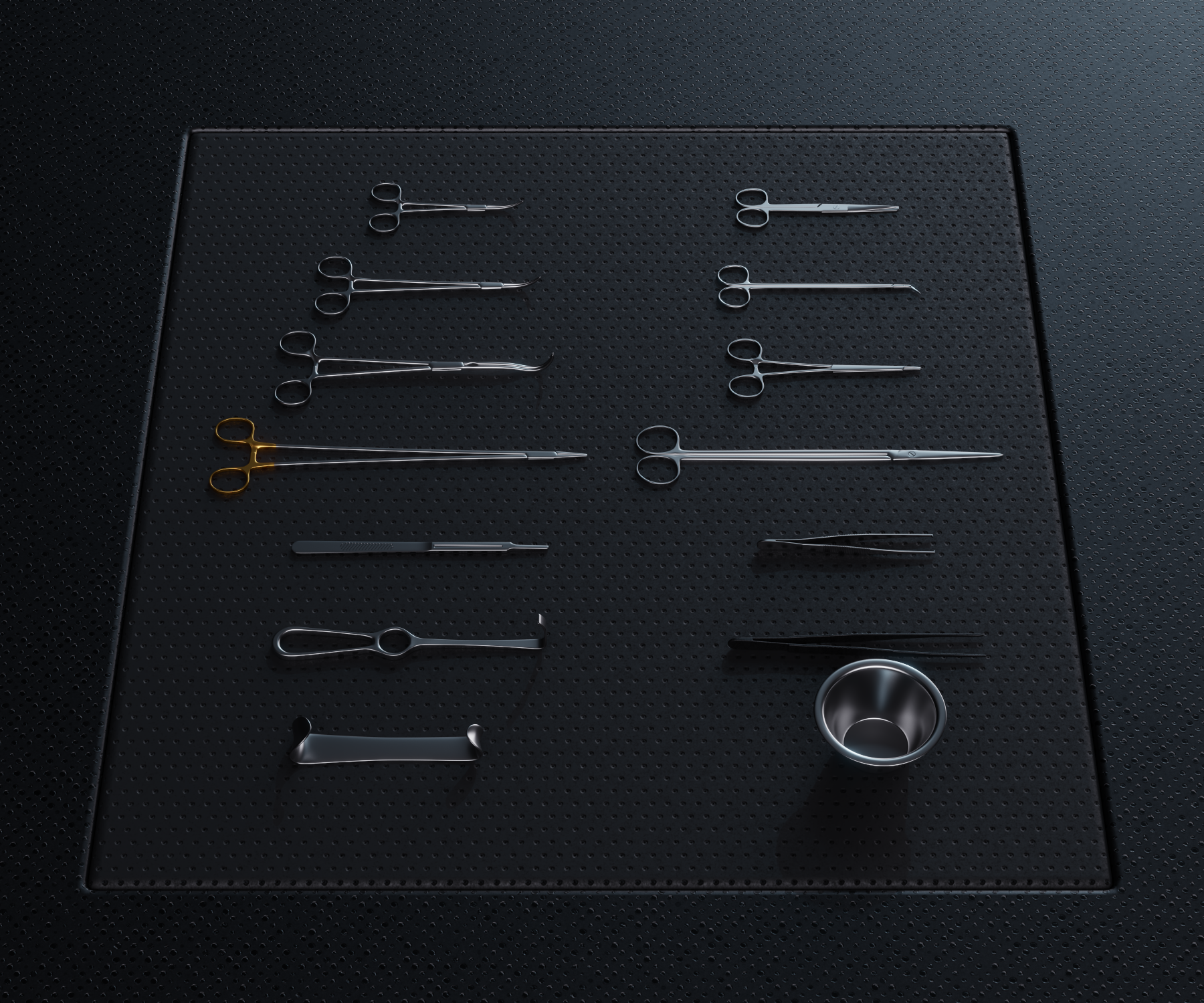}
        \caption{} 
        \label{fig:tools}
    \end{subfigure}
    \begin{subfigure}{0.58\textwidth}
        \includegraphics[width=0.99\linewidth]{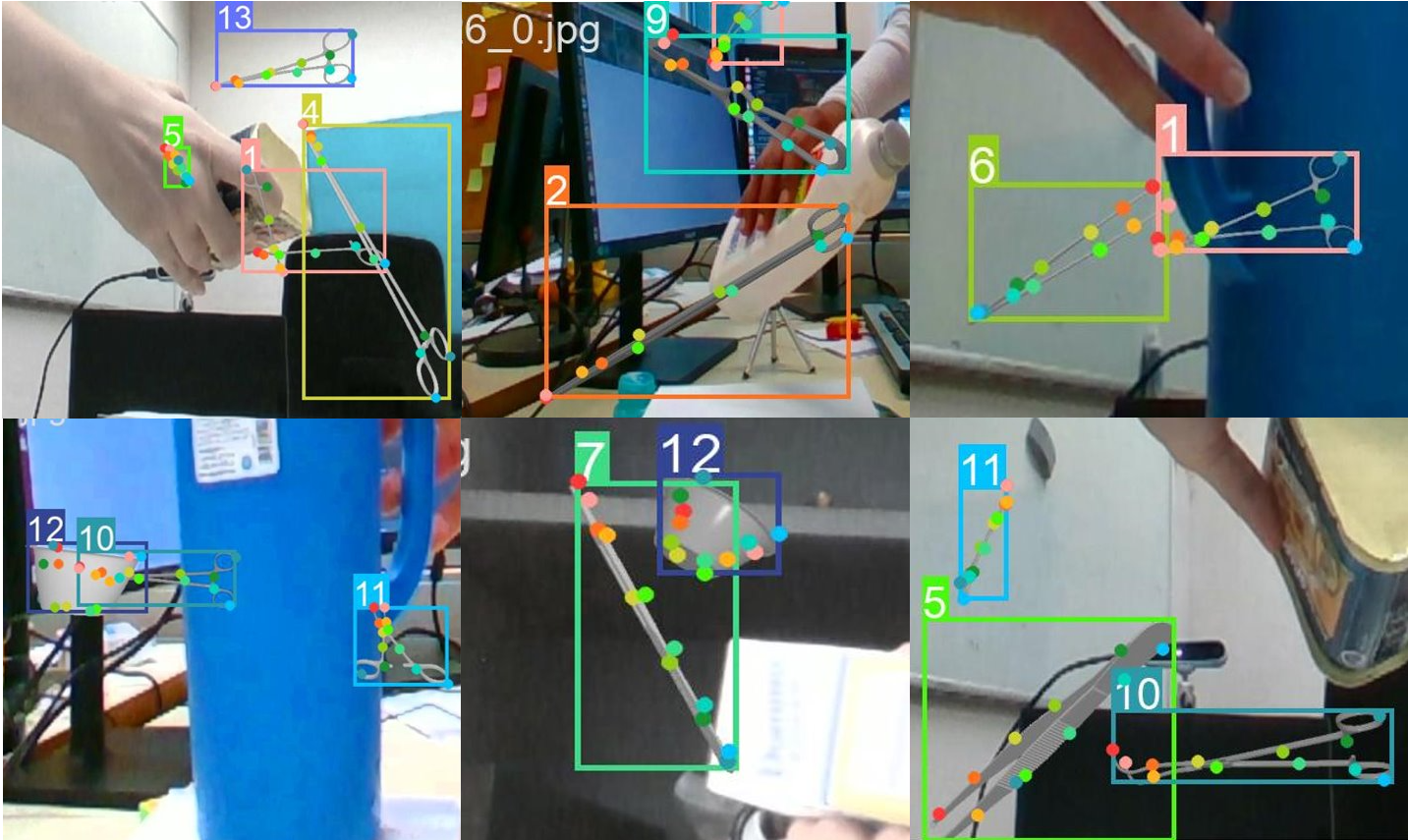}
        \caption{} 
        \label{fig:yoloSample}
    \end{subfigure}

    \caption{a) A rendered synthetic image of the studied set of surgical instruments. b) Synthetic images generated using PyTorch3D that include medical instruments in random poses with their annotations on background images from the HO-3D dataset\cite{hampali2020honnotate}. }
    \label{fig:data}
\end{figure}

To train a network for surgical instrument detection and keypoint estimation from RGB images, we use PyTorch3D~\cite{ravi2020pytorch3d} to synthesize RGB images of objects in random poses.
During the synthesis process, we select a random subset of the surgical instruments' meshes and randomize their 7D pose. 
The corresponding rendered meshes are then projected into random backgrounds.
The annotations of that image are the random 7D transformation, the $12$  keypoints along with the object classes and their bounding boxes.
The resulting total number of generated samples is around 10K images. 
Figure \ref{fig:yoloSample} shows samples of rendered images with plotted annotations.

\subsubsection{Synthetic Transformer Dataset}
\label{sec:poseDataset}
To train the Transformer network shown in Figure \ref{fig:network}, 
we follow a similar approach to the synthetic RGB dataset 
in which we randomize a pose and apply it to a mesh. 
Instead of rendering the mesh on a random background, we project it to both left and right camera coordinates to get the stereo 2D keypoints of that pose. 
The final dataset has around 100k samples of stereo 2D keypoints and their corresponding 7D poses.

\section{Experiments}

\begin{figure}
    \centering
    \includegraphics[width=\linewidth,trim={0 0.5cm 0 0},clip]{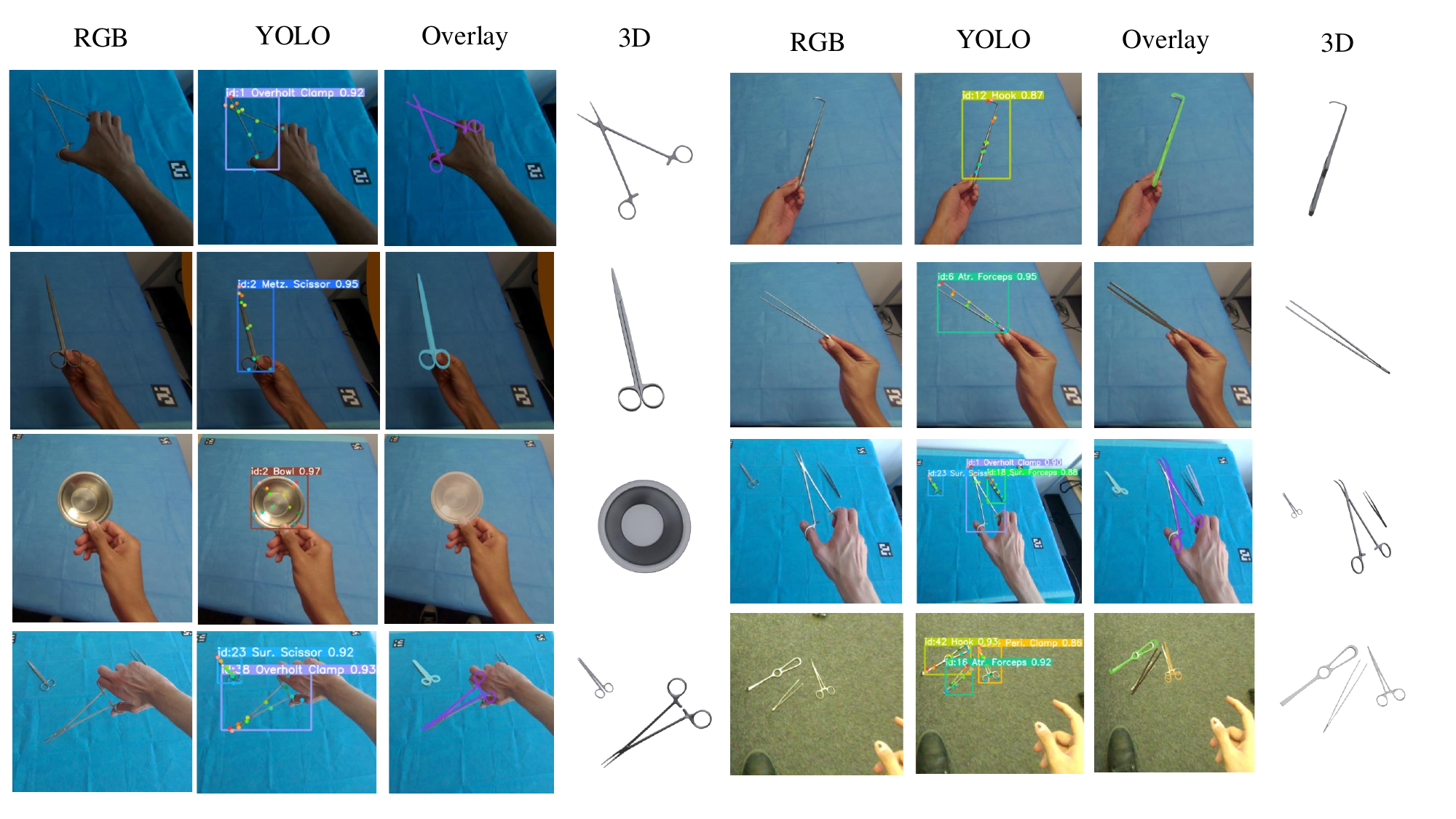}
    \caption{The results of SurgeoNet on real unseen images.}
    \label{fig:results}
\end{figure}

The main qualitative results of our approach are summarized in Figure \ref{fig:results}.
In the following sections, we quantitatively evaluate the different components of our pipeline.
We first compare the quality and performance of different YOLOv8 architectures with varying image resolutions. 
We additionally show the impact of the amount of training data on the first stage and its confusion scores regarding the studied set of objects.
Finally, we conduct an ablation study on the design of the Transformer network and compare it to an alternative optimization-based approach.

\subsection{YOLOv8 Evaluation on Surgical Instruments}

\begin{figure}
    \begin{subfigure}{0.49\textwidth}
        \includegraphics[width=0.99\linewidth]{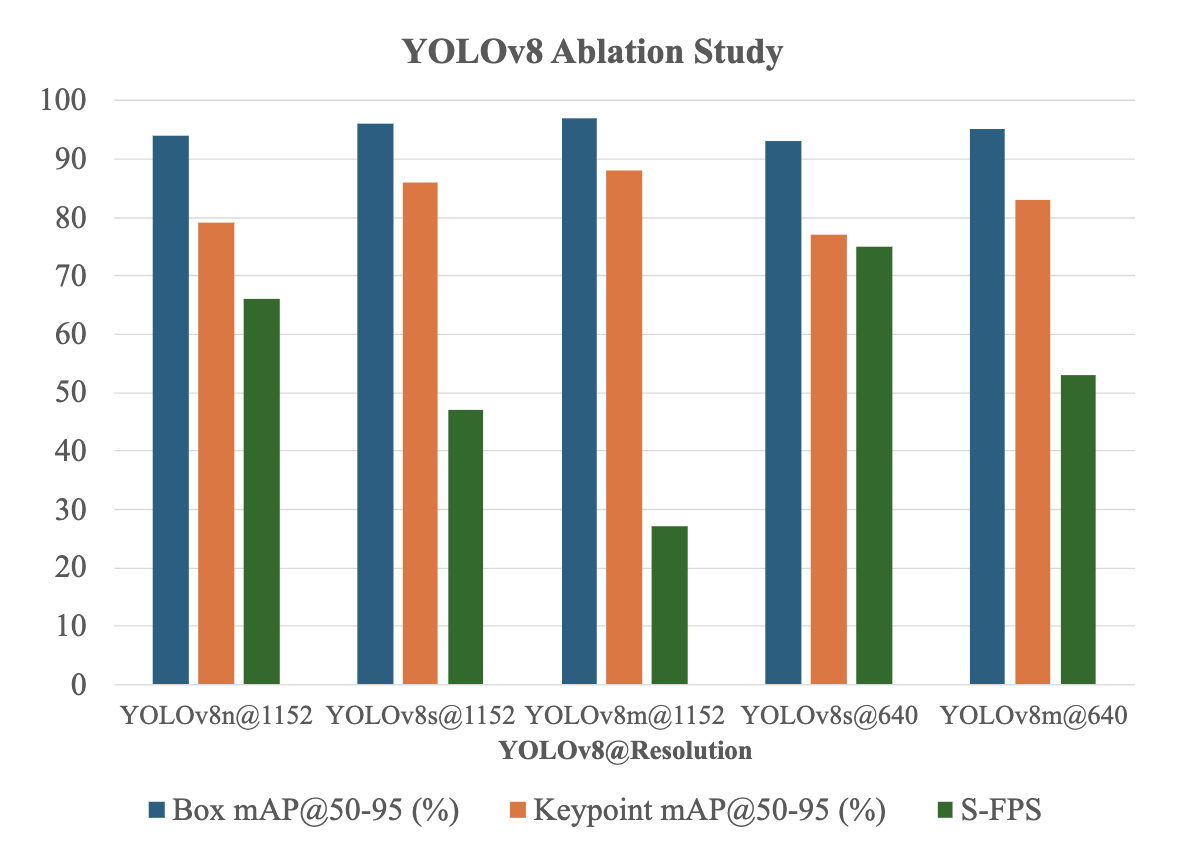}
        \caption{} 
        \label{fig:yolo_ablation}
    \end{subfigure}
    \begin{subfigure}{0.49\textwidth}
        \includegraphics[width=0.99\linewidth]{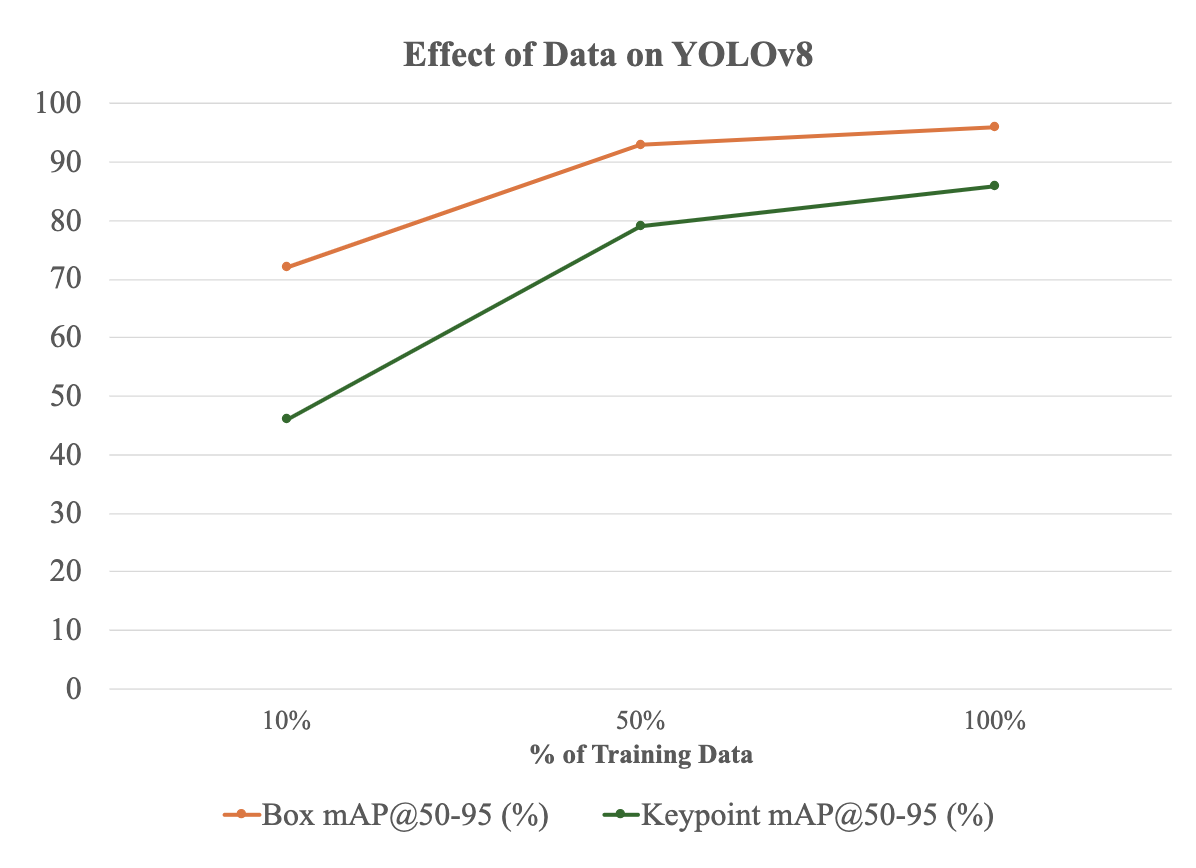}
        \caption{} 
        \label{fig:yolo_data}
    \end{subfigure}
    
    \caption{YOLOv8 Ablation Study: a) The impact of the YOLOv8 architecture and image resolution on the accuracy (Box and Keypoint mAP5@50-95) and runtime performance (S-FPS). b) The impact of the amount of training data on YOLOv8's performance.}
    \label{fig:yolo_results}
\end{figure}

\textbf{Network and Resolution}
There exist multiple YOLOv8 models that differ in the number of parameters and speed. 
To evaluate those models on our task, we finetune them on the synthetic train dataset for 200 epochs. 
We then run an evaluation on the test set to measure both the bounding box and keypoint mean Average Precision (mAP) at 50-95 IoU thresholds.
Runtime performance measured in Stereo Frames per Second (S-FPS) is calculated by running inference on a sequence of stereo RGB images captured using the Varjo and calculating the average over the sequence. 
We use the TensorRT format and run inference on an NVIDIA GeForce RTX 3090.
The results are summarized in Figure \ref{fig:yolo_ablation}.

In this work, we focus on YOLOv8m with 640 resolution and YOLOv8s with 1152 resolution as they both provide real-time performance while maintaining high accuracy.

\textbf{Required amount of training samples}
To evaluate the impact of the number of synthetic samples required for training the instruments' detection stage, we train two more networks on 10\% and 50\% of the total 8k training synthetic images. 
The results are summarized in Figure \ref{fig:yolo_data}.

\begin{figure}
    \centering
    \includegraphics[width=0.97\linewidth]{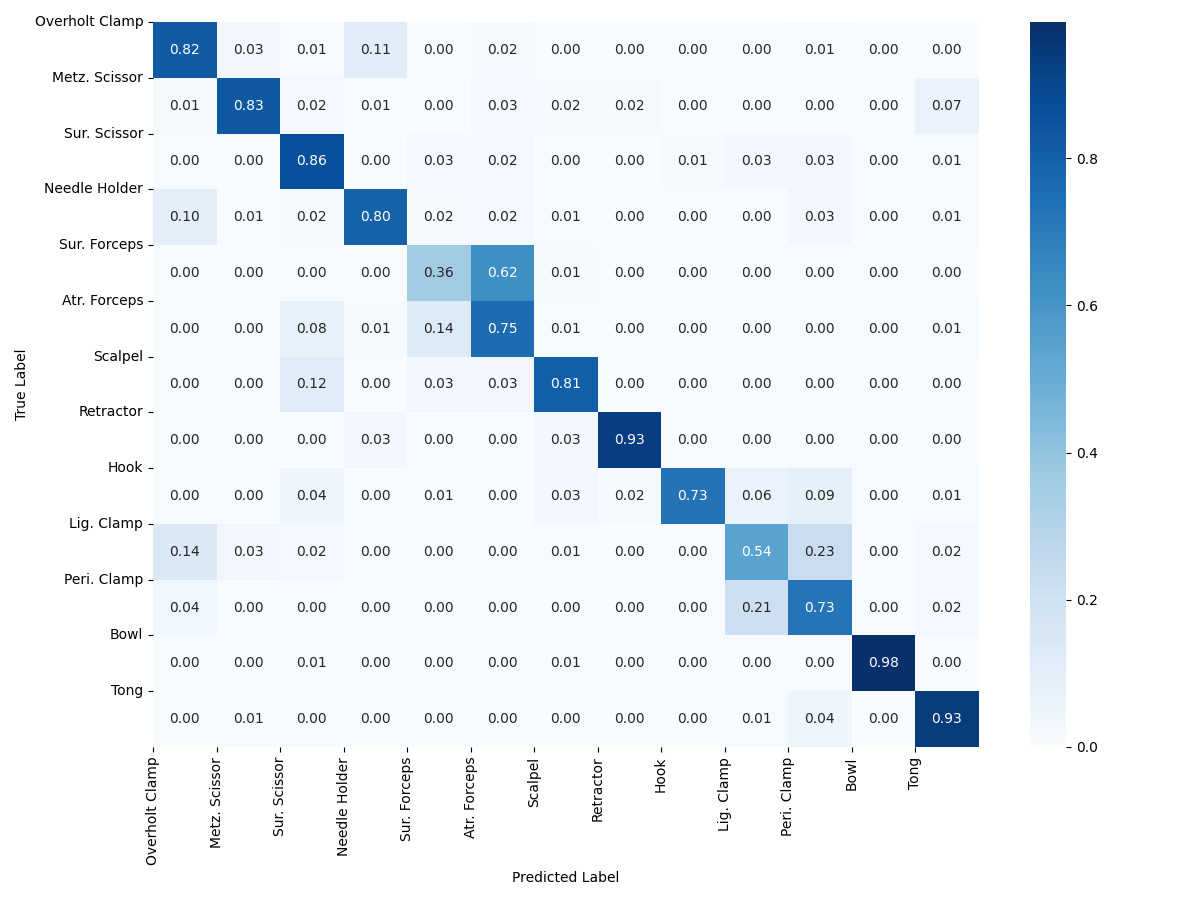}
    \caption{Confusion matrix of YOLOv8 for the 13 surgical instruments.}
    \label{fig:confusion}
\end{figure}

\textbf{YOLOv8 Confusion Matrix}
We test the classification performance of YOLO on surgical instruments and its ability to distinguish between similar objects e.g. scissors-like objects.
We record a sequence for each object in our dataset using a VR headset resulting in 13 sequences. 
Afterward, we run inference on all frames knowing the ground truth class of each frame, and compare it to YOLO's predictions.
Figure \ref{fig:confusion} shows the normalized confusion matrix. 
Except for the surgical forceps, which only differ in size, the remaining instruments are correctly classified despite their strong similarities.

\subsection{Ablation Study on the Stereo Transformer}

In this section, we show the importance of input and output formats used to design the transformer along with its hyperparameters.
We show the importance of using stereo 2D keypoints instead of using only monocular 2D keypoints. 
Furthermore, we observe the increased performance that results from using a one-hot encoding to describe each keypoint class i.e. positional embedding for keypoints.
Finally, we adopt the 6D rotation representation \cite{zhou2019continuity} instead of $3$ axis angles. 
Table \ref{tab:modality} summarizes the results of training Transformer networks on the different I/O modalities for $100$ epochs and testing them on the synthetic dataset described in Section \ref{sec:poseDataset}.
We use the Mean Per Vertex Position Error (MPVPE) in mm to describe the quality of the pose.
To study the architecture of the Transformer, we experiment with two hyperparameters, namely, the number of multi-headed attention layers and hidden dimension representation size. 
No significant impact was observed by tuning the hyperparameters and the best combination for the number of layers and hidden dimension size is $5$ and $128$ respectively.

\textbf{Transformer vs Fitting}
The final experiment is to compare the Transformer to the optimization-based keypoint fitting on two recorded real sequences.

We use an Adam optimizer~\cite{kingma2014adam} to optimize the 7D pose $\posepred$. 
During optimization, $\mathcal{T}_{wld}$ transforms the predefined set of 3D keypoints of the object $\mathcal{K}_{o}$ from their original position using the optimized pose $\posepred$. 
After that, $\mathcal{T}_{cam}$ projects the mesh to the pixel coordinates of both stereo frames using the camera matrix $\mathcal{M}_{c}$.
Finally, the loss is computed as the difference between the projected keypoints and $\kppred$ predicted by YOLO.

\begin{equation}
    \mathcal{L}_{r} = \frac{1}{2} \sum_{c} \Vert \mathcal{T}_{cam} (\transformk{\posepred}, \mathcal{M}_{c})) - \kppred \Vert_2
    \label{eq:optim}
\end{equation}

In the first frame of the sequence, we initialize the pose of the objects in the scene with $500$ iterations of optimization. 
In the subsequent frames, we use the pose of the previous frame as an initialization and limit the number of iterations to $100$. 
We also apply early stopping on the optimization process whenever the reprojection error as described in Equation \ref{eq:optim} becomes less than $4$ pixels to improve runtime while maintaining good qualitative output. 
From the results shown in Table \ref{tab:optim}, we can infer that the optimization process can sometimes be more accurate than the Transformer. 
However, optimization heavily relies on the number of iterations needed to converge which makes it very slow compared to the Transformer, and hence, not suitable for real-time applications.

\begin{table}[]
\parbox{.5\linewidth}{
    \caption{Ablation Study on the I/O Transformer Modality}
    \label{tab:modality}
    \centering
    \begin{tabular}{|c|c|c|c|c|}
    \hline
         Mono & Stereo & Kp Cls. & 6D Rot. & MPVPE (mm) \\
         \hline
         \cmark & \xmark & \xmark & \xmark & 64.0 \\ 
         \xmark & \cmark & \xmark & \xmark & 28.9 \\ 
         \xmark & \cmark & \cmark & \xmark & 23.0 \\ 
         \xmark & \cmark & \cmark & \cmark & \textbf{11.8} \\ 
         \hline
    \end{tabular}
    }
\hfill
\parbox{.45\linewidth}{

    \caption{Comparison between the Transformer method and the optimization-based fitting method.}
    \label{tab:optim}
    \begin{tabular}{|c|c|c|c|c|}
    \hline
     & \multicolumn{2}{c|}{Transformer} & \multicolumn{2}{|c|}{Optimization} \\
     \cline{2-5}
    Obj. Cls. & Err. & FPS & Err. & FPS \\
    \hline
    1 & 16.9 & 209 & 13.8 & 1 \\
    13 & 11.8 & 202 & 21.6 & 1 \\
    \hline

    \end{tabular}
    
}
\end{table}

\subsection{State-of-the-art Comparison}

To compare our method as a surgical instrument pose estimator, we train and evaluate the network on the Hein~\etal~\cite{hein2021towards} surgical drill dataset. 
In addition, we evaluate our method as a stereo-based object pose estimator on the StereOBJ-1M~\cite{liu2021stereobj} benchmark dataset.

\subsubsection{Hein~\etal~\cite{hein2021towards} (Drill)}
\label{sec:drill}
is a dataset containing real and synthetic monocular 256x256 frames of a surgical drill being used in an operation room.
In this work, we only focus on the real dataset for training and testing.
The total number of real frames is $3,746$ and we follow the same fivefold cross-validation evaluation setup mentioned by the authors.
To train the network, we sample 12 keypoints from the drill mesh and train a YOLOv8-m and the Transformer with the additional drill class.

Given the rigidity of the object, the ADD metric (Average Distance of Model Points) is used for evaluation:
\begin{equation}
\text{ADD} = \frac{1}{|M|} \sum_{x \in M} \| (Rx + t) - (\hat{R}x + \hat{t}) \|
\end{equation}
where $M$ are the mesh 3D points, $R$ and $t$ represent the ground truth pose, and $\hat{R}$ and $\hat{t}$ represent the predicted pose.
The results in Table \ref{tab:drill} suggest that the Transformer can find an accurate pose if given correct 2D keypoints from a single view. 
However, when given the YOLO predictions, the network produces lower accurate poses in comparison to previous methods.

\begin{table}
    \centering
    \caption{Average ADD error across fivefold cross-validation test sets.}
    \label{tab:drill}
    \begin{tabular}{|c|c|}
    \hline
         Model & Tool ADD (mm) \\
         \hline
         HandObjectNet~\cite{hasson2020leveraging} & $13.8$ \\
         PVNet~\cite{peng2019pvnet} & $39.7$ \\
         HMD-EgoPose~\cite{doughty2022hmd} & $17.2$ \\
         Ours (w/ perfect keypoints) & $11.4$ \\
         Ours (w/ YOLO keypoints) & $44.3$ \\
         \hline
    \end{tabular}
\end{table}

\subsubsection{StereOBJ-1M~\cite{liu2021stereobj}}

contains stereo RGB frames with a resolution of 1440x1440 along with 6-DoF annotations for all rigid objects in the scene, and predefined meshes and keypoints. 
The total number of objects in the dataset is $18$.
The dataset consists of $394,612$ stereo frames, of which $274,613$ are used for training.
To train our network, we scale the resolution of the images and keypoints to 640x640 and train a YOLOv8-m model and a Transformer for $20$ epochs each. 

\begin{figure}
    \centering
    \includegraphics[width=\linewidth,trim={0 0.5cm 0 0},clip]{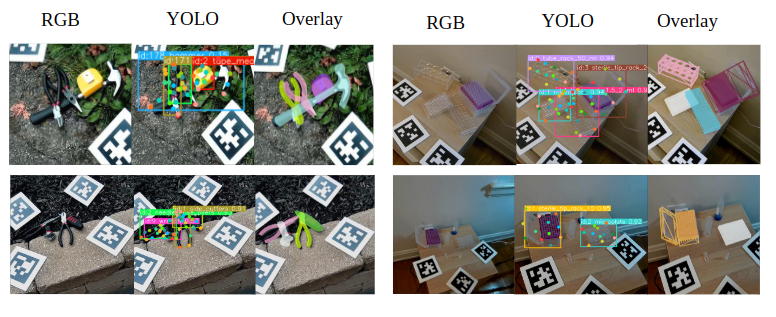}
    \caption{The results of SurgeoNet on StereOBJ-1M test set.}
    \label{fig:stereobj}
\end{figure}

To evaluate our method, we use the ADD-S Accuracy metric proposed by the dataset authors, defined: 
\begin{equation}
\text{ADD-S} = \frac{1}{|M|} \sum_{x_1 \in M} \min_{x_2 \in M} \| (Rx_1 + t) - (\hat{R}x_2 + \hat{t}) \|
\end{equation}
with the same definition of variables used in Section \ref{sec:drill}.
The ADD-S accuracy considers a pose correct if the ADD-S distance is less than 10\% of the object's diameter.
Table \ref{tab:stereobj} summarizes our results on the StereOBJ-1M benchmark dataset. 
The detailed results can be found on the challenge website\footnotemark.
Despite being the lowest on average accuracy, SurgeoNet shows competitive results on multiple objects compared to State-of-the-art achieving best scores on some of them. 
This shows improvements on multiple objects as seen in the last row of Table~\ref{tab:stereobj}.
Figure \ref{fig:stereobj} shows qualitative results on the StereOBJ-1M test set.
The results suggest that our method is robust in cluttered scenes and with transparent objects.

\begin{table}[]
    \centering
    \caption{Average ADD-S Accuracy and ADD-S Accuracy on a selected subset of objects (abbreviated with object initials) from the StereOBJ-1M benchmark dataset. 
    }\label{tab:stereobj}
    \begin{tabular}{|c|c|c|c|c|c|c|c|c|c|}
        \hline
        Model & Average & M & NNP & P100 & SC & STR200 & TR1.5 & TR50 & WS \\
        \hline
        PVNet~\cite{peng2019pvnet} & \textbf{42.48} & $18.72$ & $51.78$ & $0.65$ & \textbf{69.11} & $62.52$ & $52.05$ & $75.04$ & \textbf{72.63} \\
        KeyPose~\cite{liu2020keypose} & $39.42$ & \textbf{39.22} & $51.72$ & $1.77$ & $39.12$ & \textbf{67.04} & $60.10$ & $72.05$ & $71.78$ \\
        Ours & $36.46$ & $29.22$ & \textbf{52.01} & \textbf{6.81} & $51.40$ & $59.21$ & \textbf{72.50} & \textbf{87.12} & $62.52$\\
        \hline
    \end{tabular}
\end{table}

\footnotetext{\url{https://eval.ai/web/challenges/challenge-page/1645/leaderboard/3943}}





\section{Conclusion}
In this work, we presented SurgeoNet, a new real-time neural-network pipeline to accurately reconstruct temporally-consistent 7D poses of articulated surgical instruments from stereo VR-view. The approach builds on top of state-of-the-art architectures, including YOLO and Transformers. Thanks to its real-time performances, the approach is suitable for mixed-reality
applications, especially in medical scenarios involving hand-object interactions with surgical tools. We demonstrated the method's robustness in the classification of thin articulated surgical instruments of similar shape
and appearance in challenging settings with occlusions. As shown in the evaluation, SurgeoNet demonstrated strong generalization capabilities to real sequences, despite being trained exclusively on cheap synthetic dataset.

Future work includes handling hand-object interactions to fine-tune the predicted pose. 
In addition, long-term temporal information from sequential frames will be used to improve the model's performance.

\textbf{Acknowledgements:} 
This work was partially funded by the Federal Ministry of Education and Research of the Federal Republic of Germany (BMBF), under grant agreements: GreifbAR [Grant Nr 16SV8732], and DECODE [Grant Nr 01IW21001].

\bibliographystyle{splncs04}
\bibliography{bibliography}

\end{document}